\documentclass[letterpaper, 10 pt, conference]{ieeeconf}  

\IEEEoverridecommandlockouts                              

\usepackage{eso-pic} 
\usepackage[utf8]{inputenc} 
\usepackage[T1]{fontenc}    
\usepackage{hyperref}       
\usepackage{url}            
\usepackage{booktabs}       
\usepackage{amsfonts}       
\usepackage{amsmath}
\usepackage{nicefrac}       
\usepackage{microtype}      
\usepackage{sidecap}		
\usepackage{graphicx}

\makeatletter
\let\NAT@parse\undefined
\makeatother
\usepackage[numbers]{natbib}

\usepackage{lipsum}
\usepackage{multirow}
\usepackage{multicol}
\usepackage{xcolor,colortbl}
\usepackage{multirow}
\usepackage[font=small,labelfont=bf]{caption}
\usepackage{subcaption}
\usepackage{wrapfig}
\usepackage{footnote}
\usepackage{flushend} 
\let\labelindent\relax
\usepackage{enumitem}
\usepackage{lscape}      
\usepackage{flushend}

\makesavenoteenv{tabular}
\makesavenoteenv{table}

\definecolor{Light}{RGB}{193, 237, 246}
\definecolor{Gray1}{gray}{0.0}
\definecolor{Gray2}{gray}{0.1}
\definecolor{Gray3}{gray}{0.25}
\definecolor{Gray4}{gray}{0.4}

\usepackage[ruled,vlined]{algorithm2e}
\newcommand{\secref}[1]{Sec.~\ref{#1}}

\newcommand{\figref}[1]{Fig.~\ref{#1}}
\newcommand{\tabref}[1]{Table~\ref{#1}}

\renewcommand{\baselinestretch}{0.99}

\title{Transferring Dexterous Manipulation from GPU Simulation \\ to a Remote Real-World TriFinger}

\author{
   Arthur Allshire$^{1,2}$, 
   Mayank Mittal$^{2,3}$, 
   Varun Lodaya$^{1}$, 
   Viktor Makoviychuk$^{2}$, 
   Denys Makoviichuk$^{4}$, \\
   Felix Widmaier$^{5}$, 
   Manuel Wüthrich$^{5}$, 
   Stefan Bauer$^{6}$, 
   Ankur Handa$^{2}$, 
   Animesh Garg$^{1,2}$%
\thanks{$^{1}$University of Toronto, Vector Institute,
   $^{2}$Nvidia,
   $^{3}$ETH Zurich,
   $^{4}$Snap,
   $^{5}$MPI Tubingen,
   $^{6}$KTH.
   Email: \texttt{arthur@allshire.org}
   }
}

\usepackage{float}

\begin{document}

\makeatletter
    \let\@oldmaketitle\@maketitle
    \renewcommand{\@maketitle}{\@oldmaketitle
    \begin{minipage}[c]{\textwidth}
    \centering
    \includegraphics[width=0.9\textwidth]{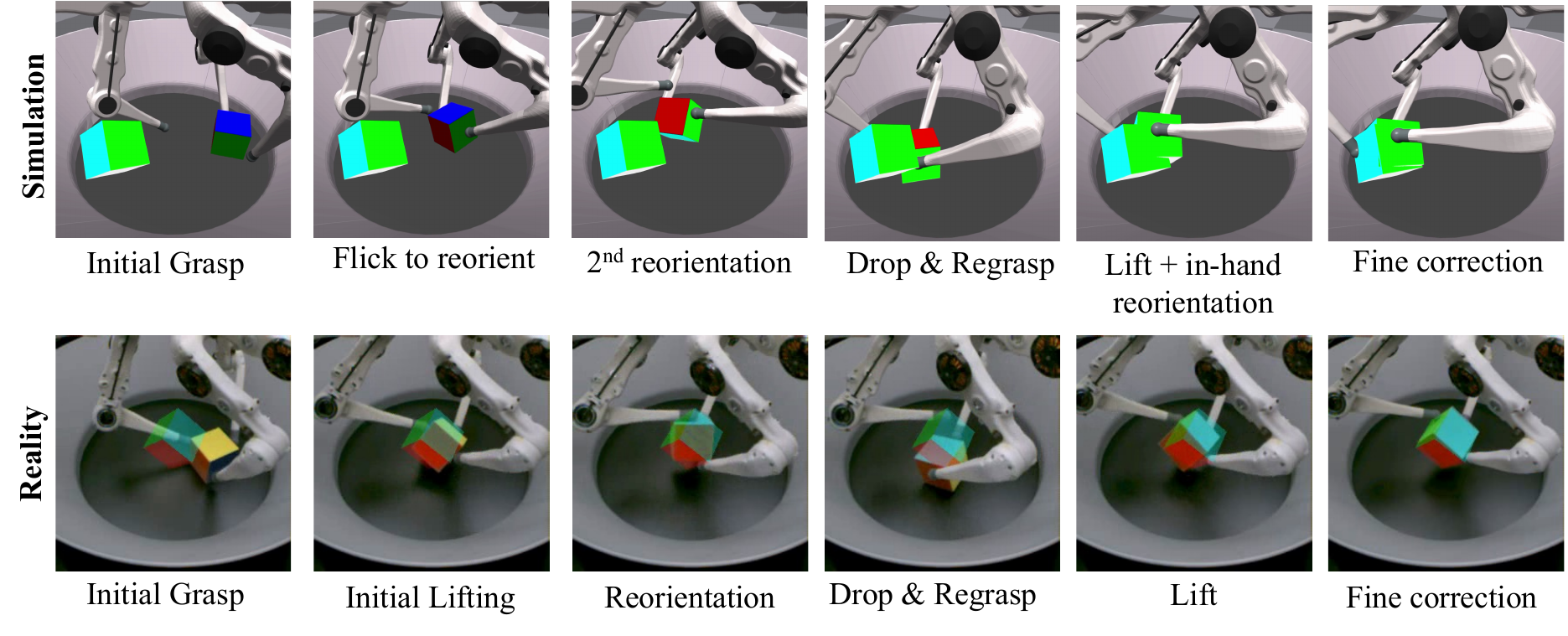}
    \captionof{figure}{\textit{Top:} Our system learns to grasp and manipulate objects to 6-DoF goal poses with a single policy, entirely in simulation, across a variety of objects. \textit{Bottom:} We then transfer to a real robot located thousands of kilometers away from where development work is done.}
    \label{fig:teaser}
   \end{minipage}
  \vspace{-15pt}
    }
\makeatother

\maketitle
\IEEEpeerreviewmaketitle

\begin{abstract}
In-hand manipulation of objects is an important capability to enable robots to carry-out tasks which demand high levels of dexterity. This work presents a robot systems approach to learning dexterous manipulation tasks involving moving objects to arbitrary 6-DoF poses. We show empirical benefits, both in simulation and sim-to-real transfer, of using keypoint-based representations for object pose in policy observations and reward calculation to train a model-free reinforcement learning agent. By utilizing domain randomization strategies and large-scale training, we achieve a high success rate of 83\% on a real TriFinger system, with a single policy able to perform grasping, ungrasping, and finger gaiting in order to achieve arbitrary poses within the workspace. We demonstrate that our policy can generalise to unseen objects, and success rates can be further improved through finetuning. With the aim of assisting further research in learning in-hand manipulation, we provide a detailed exposition of our system and make the codebase of our system available, along with checkpoints trained on billions of steps of experience, at \url{https://s2r2-ig.github.io}
\end{abstract}


\section{Introduction}

%

Multi-fingered robotic platforms are essential for executing complicated tasks such as fruit harvesting, and circuit assembly. However, performing such dexterous manipulation tasks requires dealing with various challenging factors. These include high-dimensionality, hybrid dynamics, and uncertainties about the environment~\cite{okamura2000overview}.

Therefore, performing multi-fingered manipulation in the context of deployed systems poses unique challenges which practical controllers must overcome. In our work, we provide a method for creating controllers capable of achieving reliable dexterous manipulation tasks across a variety of domain shifts, both from sim to real and across different simulation scenarios. We demonstrate the robustness of our system through: showing transfer of policies created in simulation to the real world, varying system parameters and showing robustness, and changing the object being manipulated.

Our controllers are trained and evaluated on the TriFinger~\citep{trifinger-platform} hand. Despite the challenging configuration of the Trifinger system (hand oriented down) and task (grasping and subsequent reposing), we train a unified  neural-network to achieve effective and robust control over the system. 

The Trifinger robots are run as cloud-based robot farms. Such remote robotic systems promise to alleviate many of the upfront requirements to install and maintain robot hardware~\cite{kehoe2015cloudrobsurvey}. However, RaaS systems are also more rigid than commonly used research platforms, as controllers designed for individual platforms must be rolled out across the entire fleet. Hence, the robustness of our learning-based approach is further proven out by deployment on a system which we lacked physical access to.


The key insight of this paper lies in a careful integration and evaluation of empirical advances in reinforcement learning with high-speed simulation, and practical deployment on a remote robot system. 
In particular, contribution of this robot systems work are:
\begin{enumerate}[noitemsep,topsep=0pt,leftmargin=1em]
    \item We provide a framework for learning the skill of in-hand manipulation tasks robust to sim-to-real transfer and object morphology.
    \item Unlike previous in-hand manipulation systems, we produce a single policy which performs both grasping and re-posing, simplifying the pipeline of using RL in such systems.
    \item  We show the benefits of using keypoints to represent the object in RL algorithms for in-hand manipulation, especially when reposing in 6-DoF.
    \item We show our system is robust to changing object morphology and physics parameters, shown through simulated experiments and demonstrated sim-to-real transfer.
    
\end{enumerate}




\vspace{-3pt}
\section{Related Work}
\vspace{-3pt}

\subsection{Learning Dexterous Manipulation with Robot Hands}

\noindent \textbf{1. Reinforcement Learning} Advances in RL algorithms and computational hardware have enabled rapid progression in the capability of real robots in dynamic scenes. Techniques such as domain randomization and large-scale training have enabled results across a variety of tasks with sim-to-real, including in-hand manipulation \citep{openai-sh, openai-rubiks}, as well as in legged locomotion \citep{Hwangbo_2019, shi2020circus}. Active identification of system parameters has also been shown to be helpful in the context of learning manipulation tasks~\citep{chebotar2019closing}.

\noindent \textbf{2. Dextrous Manipulation} Dexterous manipulation requires dealing with high-dimensionality of the system, hybrid dynamics, and uncertainties about the environment~\cite{okamura2000overview}. Prior work has trained a control policy for in-hand manipulation of a block with a Shadow Dexterous Hand~\cite{openai-sh}. However, this relied on expensive robot hardware in a controlled environment which could be reset and tuned by engineers.  Recent works have extended this setup to other object morphologies \citep{huang21, chen21}.

\noindent \textbf{3. Hand Platforms} Recently,~\citet{trifinger-platform} designed an open-sourced robotic platform for dexterous manipulation called \emph{TriFinger}. The Trifinger platform allows remote deployment of policies on a pre-determined experimental setup \citep{reproducible-cluster}, without tweaking of system parameters. As a result of the "reset-free" nature of the robot, the hand is facing down, and objects must first be picked up from the ground to be manipulated. Prior works do not provide a method to learn this behaviour in a simple way, either choosing to address only one step in this process or resorting to chaining multiple policies together to accomplish the complete reorientation task \citep{openai-sh, chen21}. In contrast, in this work, we present an approach to address these shortcomings.

Our system is able to perform 6-DoF in-hand manipulation, as opposed to just 3-DoF reorientation. Furthermore, as the object starts outside of the hand, our single learned policy is able to perform not just picking, but also grasping and un-grasping, in contrast to these prior works.

\vspace{-3pt}
\subsection{Sim-to-real transfer with RL Policies}
\vspace{-3pt}

\noindent \textbf{1. Challenges of Sim-to-Real} Our method relies on learning control policies using gradient-based optimisation combined with large-scale accelerated simulation, a proven method of learning a wide variety of complex robotic tasks \citep{makoviychuk2021isaac, brax2021github, mandlekar2017arpl, Jason:ICRA:2018}. However, doing so means that policies must solve the "sim-to-real transfer" problem of being robust to inference in a different environment than which they were trained. Sim-to-real transfer of manipulation policies is a challenging problem for two major reasons: 1) differences between real-world environment interactions and that of the simulation where policies are trained, and 2) state estimation of objects being manipulated. For the former, a variety of practical methods have been proposed including Domain Randomisation in simulation \citep{Jason:ICRA:2018, mandlekar2017arpl}, Bayesian optimisation on the real system \citep{trifinger-benchmarking}, and optimisation of the simulator parameters \citep{chebotar2019closing, bayessim}.

\noindent \textbf{2. State Estimation} Previous methods of state estimation have included pose estimation \citep{openai-sh, tremblay2018deep} more recently, policy distillation as a method to solve sim-to-real problem for state estimation \citep{chen21, Hwangbo_2019}. However this is limited by the visual fidelity of current simulators and has only been shown to work for individual objects, limiting the generality of the policy. As a result, practical approaches leveraging existing work in vision still must rely on using pose tracking in the real world.

\setcounter{figure}{1}
\begin{figure}[!t]
\centering
\begin{subfigure}{.50\linewidth}
  \centering
  \includegraphics[trim={0.7 1.5cm 0.4 1.5cm},clip,width=.85\linewidth]{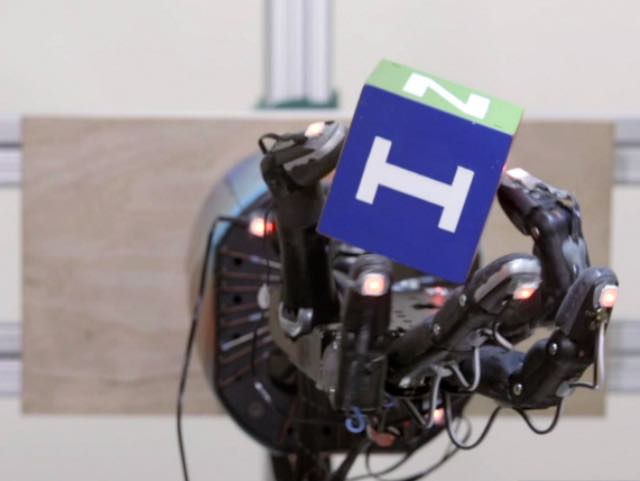}
  \caption{OpenAI's shadow hand setup. The cube starts placed in the hand.}
  \vspace{3mm}
\end{subfigure}%
\hfill
\begin{subfigure}{.46\linewidth}
  \centering
  \includegraphics[width=.9\linewidth]{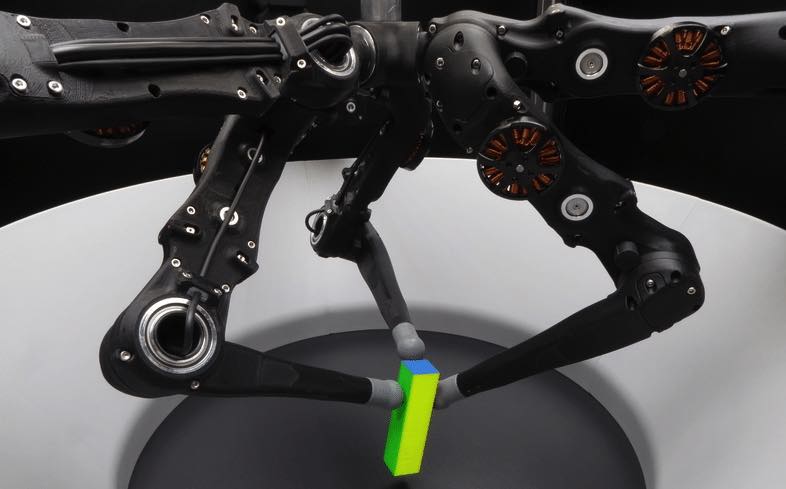}
  \caption{The Trifinger setup. The object starts outside of the hand to enable reset-free setup.}
\end{subfigure}
\caption{Previous setups for performing RL-based dexterous manipulation in the real world have relied on specialised hardware or configurations which may be impractical to scale. For example, OpenAI's work on Shadow Hand \citep{openai-sh} started with the cube in hand (avoiding the need to learn to grasp it), relied on phase space tracking, and only set in-hand orientation goals (rather than full pose goals). In contrast, the Trifinger setup relies only on sensor inputs from RGB cameras and encoders in the fingers, and the object starts in a random position on the ground outside of the hand, yet our system achieves 6-DoF reposing on multiple objects across the workspace.}
\label{fig:comparison-shadow-trifinger}
\end{figure}

\begin{figure*}[ht]
\begin{minipage}[c]{0.75\textwidth}
  \centering
  \includegraphics[width=\linewidth]{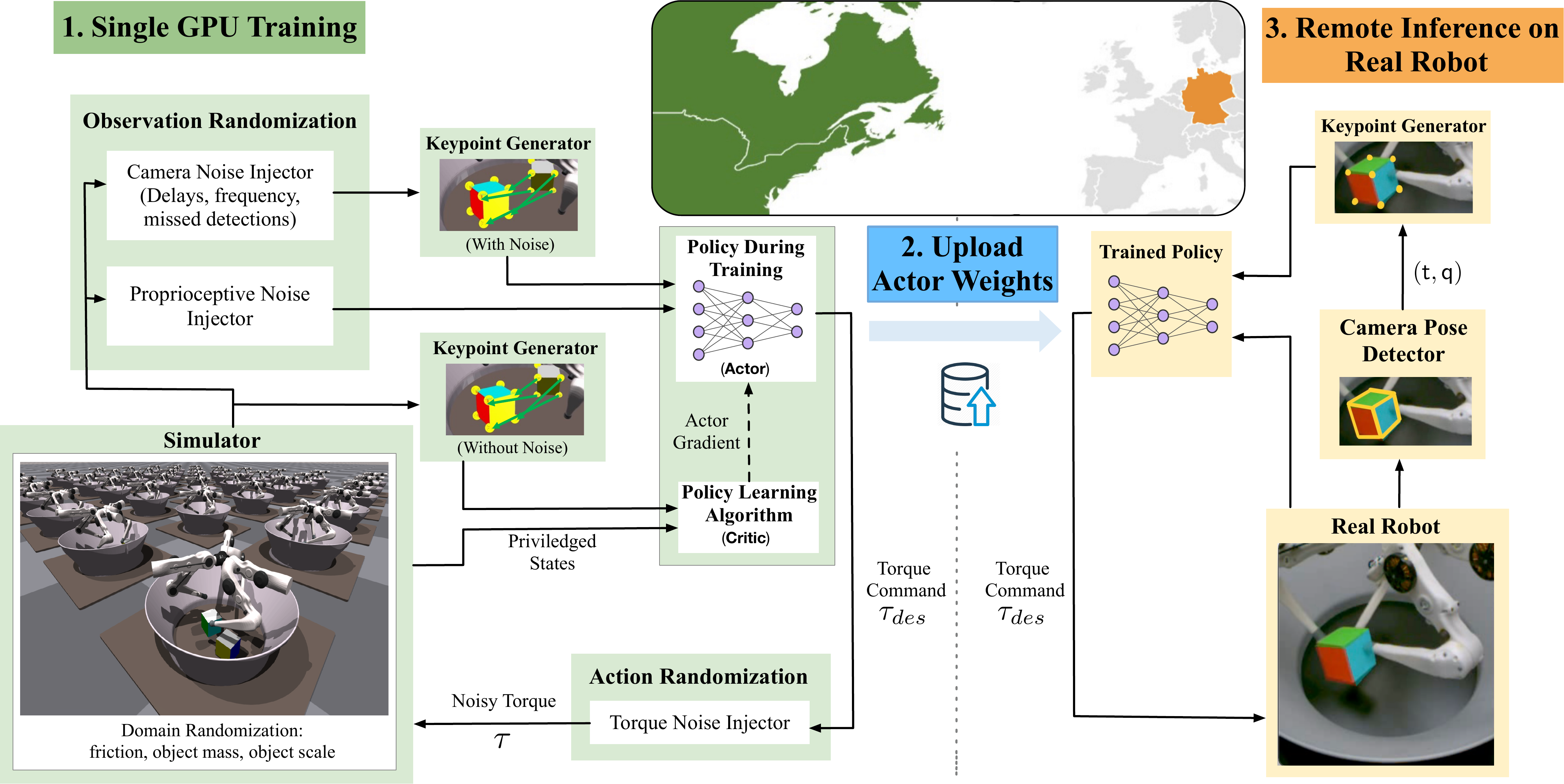}
\end{minipage}
\,
\begin{minipage}[c]{0.23\textwidth}
  \caption{Our system trains using the IsaacGym simulator\protect\footnotemark \citep{makoviychuk2021isaac} on 16,384 environments in parallel on a single NVIDIA Tesla V100 or RTX 3090 GPU. Inference is then conducted remotely on a TriFinger robot located across the Atlantic in Germany using the uploaded actor weights. The infrastructure on which we perform sim-to-real transfer is provided courtesy of the organisers of the Real Robot Challenge \citep{real-robot-challenge}.} 
  \label{fig:trifinger}
  \end{minipage}
  \vspace*{10pt}
\end{figure*}

\vspace{-3pt}
\section{Method}
\vspace{-3pt}

\subsection{Problem Setup: The Trifinger Task}
\label{sec:environment}
\vspace{-3pt}


\noindent \textbf{1. Task Description}.
In this paper, we propose a method for training a controller for the Trifinger hand \citep{trifinger-platform} to 
perform 6-DoF manipulation of objects. The objects start on the ground, and the goal of the system is to move the object to a target position and orientation and hold it there. Any solution must be able to move the fingers to the object, grasp it, and perform appropriate un-grasping and finger-gaiting to achieve the corresponding target orientation.

Our aim is to use Reinforcement Learning (RL) to learn a single policy with a unified reward model to achieve closed-loop controller for this task, in contrast to previous works which have used relied on carefully designed state machines on top of RL based low-level skills to produce such multi-modal behaviour (see \figref{fig:comparison-shadow-trifinger}). 

\noindent \textbf{2. Policy Inference on a Remote Real Robot}. It is important to show performance on real world rather than simply simulated systems. This is because of the inherent difficulty in having a simulator with the same physics parameters as a real robot, and the necessity of state estimation in the real world, making it the true test of practical methods in robotics.

On the Trifinger system, the pose of the object is tracked on the system using 3 cameras \citep{trifinger-object-tracking}. We convert the position+quaternion representation output by this system into the keypoint representation described in \secref{sec:poserepr} and use it as input to the policy. Observations of the object pose from the camera system are provided at 10Hz, compared to the higher frequency of 50Hz that the policy is run at. This means our method has to deal with relatively low-frequency and noisy object observations based on camera sensing. The torque on each joint is limited such that it does not damage the equipment while in operation, however the exact values of these parameters may vary and are not exposed, meaning our system must be robust to a range of these parameters. The policy is uploaded to the remote system and run on a the robot's local computer to mitigate latency issues.

\vspace{-3pt}
\subsection{Reinforcement Learning for Dexterous Manipulation}
\label{sec:learning}


\begin{table}[]
\centering
    \begin{subtable}[h]{\linewidth}
        \centering
        \resizebox{\linewidth}{!}{%
        \begin{sc}
        \begin{tabular}{l|l|c} 
            \toprule
            \rowcolor[HTML]{CBCEFB}
            \multicolumn{2}{c|}{Observation space}                                   & Degrees of freedom  \\ 
            \midrule
            \rule{0pt}{2ex}    
            \multirow{2}{*}{Finger joints}                        & position  & $3 \text{ fingers} \cdot 3 \text{ joints} \cdot 1[\mathbb{R}^1]=9$\\ 
                                                                  & velocity  & $3 \text{ fingers} \cdot 3 \text{ joints} \cdot 1[\mathbb{R}^1] = 9$ \\ 
            \rowcolor[HTML]{EFEFEF} 
            \multirow{1}{*}{Object pose}                           & keypoints      & $8 \text{ keypoints} \cdot3 \text{ } [\mathbb{R}^3]=24$                    \\ 
            \multirow{1}{*}{Goal pose}                           & keypoints      & $8 \text{ keypoints} \cdot3 \text{ } [\mathbb{R}^3]=24$                    \\ 
            \rowcolor[HTML]{EFEFEF} 
            \multirow{1}{*}{Last action}                        & torque  & $3 \text{ fingers} \cdot 3 \text{ joints} \cdot 1[\mathbb{R}^1]=9$ \\ 
            \midrule
            \rowcolor[HTML]{CBCEFB}
            \multicolumn{2}{c|}{Total}                                  & 75                 \\
            \bottomrule
        \end{tabular}
        \end{sc}
        }
        \caption{Actor Observations}
        \label{table:policyobs}
    \end{subtable}
    ~
    \begin{subtable}[h]{\linewidth}
        \centering
        \resizebox{\linewidth}{!}{%
        \begin{sc}
        \begin{tabular}{l|l|c} 
            \toprule
            \rowcolor[HTML]{CBCEFB}
            \multicolumn{2}{c|}{Observation space}                                   & Degrees of freedom  \\ 
            \midrule
            \multicolumn{2}{c|}{Actor Observations (w/o DR)} & 75 \\
            \rowcolor[HTML]{EFEFEF} 
            Object                        & velocity & $6 \text{ }[\mathbb{R}^6]$ \\
            \multirow{3}{*}{Fingertips state}                     & pose & $3 \text{ fingers} \cdot7 \text{ } [\mathbb{R}^3 \times SO(3)]=21$ \\ 
                                                                  & velocity & $3 \text{ fingers} \cdot6 \text{ }[\mathbb{R}^6]=18$ \\ 
                                                                  & wrench& $3 \text{ fingers} \cdot6 \text{ } [\mathbb{R}^6]=18$ \\ 
            \rowcolor[HTML]{EFEFEF} 
            \multirow{1}{*}{Finger joints}                        & torque & $3 \text{ fingers} \cdot 3 \text{ joints} \cdot 1[\mathbb{R}^1]=9$\\ 
            \midrule
            \rowcolor[HTML]{CBCEFB}
            \multicolumn{2}{c|}{Total}                                  & 147                 \\
            \bottomrule
            \end{tabular}
            \end{sc}
        }
        \caption{Critic Observations}
        \label{table:valueobs}
     \end{subtable}
     \caption{\textbf{Asymmetric actor-critic to learn dexterous manipulation.} While the actor receives noisy observations, which are added as a part of DR (\secref{sec:environment}), the critic receives the same information without any noise and also has access to certain privileged information from simulator.}
     \label{tab:temps}
\end{table}

\noindent \textbf{1. RL Formulation} We model our problem using a sequential decision making formulation in which the robotic agent interacts with the environment with the objective of maximising the sum of discounted rewards. This is modelled as a discrete time, partially observable Markov Decision Process (POMDP), represented as the tuple $(\mathcal S, \mathcal O, \mathcal A, P, r, \gamma, \mathcal S_0)$, where $\mathcal S$ is the state space, $\mathcal O$ are the observations corresponding to partial information about the system states, $\mathcal A$ is the action space,  $P: \mathcal S \times \mathcal S \times \mathcal A \to \mathbb{R}$ are the probabilistic state transition dynamics, $r$ is the reward, $\gamma$ is the discount factor per timestep, and $\mathcal S_0: S \to \mathbb{R}$ is the distribution over the system's initial state at the beginning of an episode.

\begin{figure}[t]
    \centering
    \includegraphics[width=0.98\linewidth]{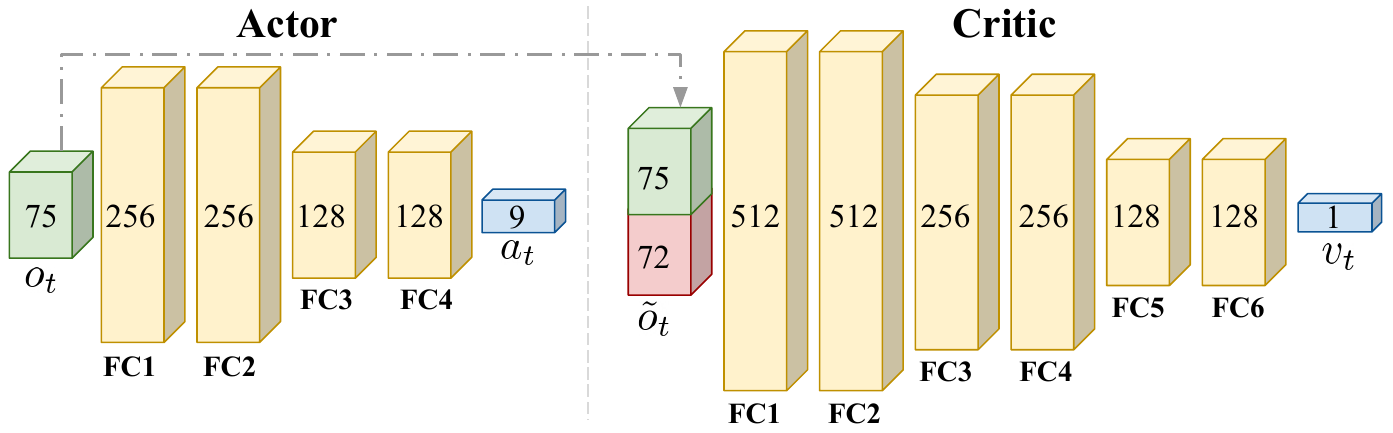}
    \caption{
    The actor and critic networks are parameterized using fully-connected layers with \texttt{ELU} activation functions~\cite{clevert2015fast}.
    }   
    \label{fig:architectures}
\end{figure}

\noindent \textbf{1. Proximal Policy Optimisation} (PPO) is an actor-critic RL algorithm \citep{schulman2017proximal} that we build on learning a parametric stochastic policy, $\pi_\theta(a, o)$ mapping from observations to an action distribution to maximise the sum of discounted rewards in each episode. Along with the policy $\pi$, PPO learns a value function $V^{\pi}_\phi(s)$ which approximates the on-policy value function. Following \citep{asymmetric-ac}, the learned value function is a function of states $\mathcal S$ rather than observations $\mathcal O$, which improves the accuracy of value function estimates. The observations $o \in \mathcal O$ of the policy are described in~\tabref{table:policyobs} and the states $s \in S$ provided in the value function are described in~\tabref{table:valueobs}.

\noindent \textbf{2. Parametrization} While multiple formulations of reward and observations are possible, we choose to use a parametrisation based on a keypoint formulation to represent the object pose, which we find boosts the ability of the RL algorithm to learn the task at hand (see Sec. \ref{sec:poserepr}, \ref{sec:reward}, \ref{sec:posereprexp}).


\noindent \textbf{3. Hyper-parameters} For PPO, we use the following hyper-parameters: discount factor $\gamma=0.99$, clipping $\epsilon=0.2$. The learning rate is annealed linearly over the course of training from $5\mathrm{e}{-3}$ to $1\mathrm{e}{-6}$. Our policy $\pi_\theta: \mathcal S \to \mathcal A$ is a Multilayer perceptron (MLP) with 4 hidden layers, 2 of size 256 followed by 2 of size 128, and 9 outputs which are scaled to the torque ranges of the real robot. Our value function  $V^\pi_\phi: \mathcal S \to R$ is an MLP with 2 layers of size 512, followed by 2 layers of size 256 and 128 each and produces a scalar value function as output. The action space $\mathcal A$ of our policy is torque on each of the 9 joints of the robot. 

\noindent \textbf{4. RL Library} We build on the implementation of PPO from the RL Games library \citep{rl-games}, which vectorizes observations and actions on GPU allowing us to take advantage of the parallelising provided by the simulator (see \secref{sec:environment}) by reducing the overhead in CPU-GPU communication typical to most CPU-based simulators. 

\vspace{-6pt}
\subsection{Representation of the Object Pose: Keypoints}
\label{sec:poserepr}

We focus on the task of manipulating an object in 6 degrees of freedom. As such, we must represent the pose of the object at multiple stages of our training pipeline. In order to capture both position and orientation in the same space in our representation, we use eight keypoints at the edges of the oriented bounding box of the object being manipulated. In the object's local frame these are denoted $\mathsf{k^L_i} \in \mathbb{R}^3, \, i=1, \dots, 8$. The keypoints in the world frame are related to those in the local frame by a transformation $\mathsf{k^C_i} = \mathbf{T} \mathsf{k^L_i}$, where $\mathbf{T}$ depends on the current pose of the object.

In \secref{sec:posereprexp}, we contrast keypoint representations to a position+quaterinon formulation used in \citep{openai-sh, pmlr-v87-liang18a}, finding that keypoints improve the policy's success rate substantially. During policy inference in the real world, we note as long as we are able to get the bounding box of the object (via classical trackers as on the Trifinger \citep{trifinger-object-tracking}, or with learning-based setups which commonly rely on the same keypoints on the oriented bounding box, such as~\citep{tremblay2018deep, cosypose}), we are able to obtain the keypoints on the object bounding box required to use this representation in policy input.



\begin{table}[!t]
    \begin{subtable}[h]{\linewidth}
        \resizebox{\linewidth}{!}{%
        \centering
        \begin{sc}
        \begin{tabular}{l|c|c|c} 
            \toprule
            \rowcolor[HTML]{CBCEFB}
            Parameter                                   &  Range & $\sigma$  & $\sigma_{corr}$ \\ 
            \midrule
            \multicolumn{4}{l}{\textbf{Observation Noise}} \\
            \rowcolor[HTML]{EFEFEF} 
            \hspace*{0.5cm} Object Position\footnotemark & [-0.30, 0.30] & 0.002 & 0.000 \\
            \hspace*{0.5cm} Object Orientation\footnotemark[\value{footnote}] \footnotetext{The noise to keypoints is not applied directly, instead it is added to the object pose in the world frame before computing the keypoints through it.} & [-1.00, 1.00] & 0.020 & 0.000 \\
            \rowcolor[HTML]{EFEFEF} 
            \hspace*{0.5cm} Finger Joint Position & [-2.70, 1.57] & 0.003 & 0.004 \\
            \hspace*{0.5cm} Finger Joint Velocity & [-10.00, 10.0] & 0.003 & 0.004 \\
            \midrule
             \multicolumn{4}{l}{\textbf{Action Noise}} \\
            \rowcolor[HTML]{EFEFEF} 
            \hspace*{0.5cm} Applied Joint Torque & $[-0.36, 0.36]$ & 0.02 & 0.01 \\
            \bottomrule
        \end{tabular}
        \end{sc}
        }
        \label{table:noise}
    \end{subtable}
    ~
    \begin{subtable}[h]{0.8\linewidth}
        \resizebox{\linewidth}{!}{%
        \centering
        \begin{sc}
        \begin{tabular}{l|c} 
        
            \toprule
            \rowcolor[HTML]{CBCEFB}
            Parameter & Scaling Distribution \\
            \midrule
            \multicolumn{2}{l}{\textbf{Environment Parameters}} \\
            \rowcolor[HTML]{EFEFEF} 
            \hspace*{0.5cm} Object Scale & $ \textrm{uniform}(0.97, 1.03)$ \\
            \hspace*{0.5cm} Object Mass & $ \textrm{uniform}(0.70, 1.30)$ \\
            \rowcolor[HTML]{EFEFEF} 
            \hspace*{0.5cm} Object Friction & $ \textrm{uniform}(0.70, 1.30)$ \\
            \hspace*{0.5cm} Table Friction & $ \textrm{uniform}(0.50, 1.50)$ \\
            \rowcolor[HTML]{EFEFEF} 
            \hspace*{0.5cm} External Forces & Refer to~\cite[pp.~9]{openai-sh} \\
            \bottomrule
            \end{tabular}
            \end{sc}
        }
     \end{subtable}
     \vspace{2pt}
     \caption{For observations and actions, $\sigma$ and $\sigma_{corr}$ are the standard deviation of additive gaussian noise sampled every timestep and at the start of each episode, respectively.  For environment, the parameters represent scaling factor applied to the nominal values in the real robot model.
     }
     \label{tab:dr}
\end{table}

\subsection{Reward Formulation \& Curriculum}
\label{sec:reward}
\noindent \textbf{1. Kernel} Our reward function $r: \mathcal S \times \mathcal A \to R$ has three components. Following \citep{Hwangbo_2019}, we use a logistic kernel to convert tracking error in Euclidean space into a bounded reward function. We generalise the kernel formulation to account for a range of distance scales, defining, $\mathcal{K}(x) = \left(e^{ax} + b + e^{-ax}\right)^{-1}$, where $a$ is a scaling factor and $b$ controls the sensitivity of the kernel at low distances.

 \noindent \textbf{2. Object Displacement Reward} As noted in \secref{sec:poserepr}, we use keypoints in order to calculate the reward in a natural space for 3-D reposing. The component of the reward corresponding to the difference between the object's current pose and the desired target pose is given by 
 $r_o = \sum_{i=1}^N \mathcal{K}(|| \mathsf{k^C_i} - \mathsf{k^T_i} ||)$, where the $\mathsf{k^C_i}$ and $\mathsf{k^T_i}$ lie at the $N=8$ vertices of the bounding boxes of the object at the current and target configurations respectively (see \secref{sec:poserepr}).
 

 \noindent \textbf{3. Finger Reaching Reward} To encourage the fingers to reach the object during initial exploration, we give a reward for moving the fingers towards the object, which was also found to be helpful in \citep{causalworld}. This term is defined by sum of the movement of each fingertip towards the goal per timestep: $r_f = \sum_{i=1}^3 \Delta^t_i$, where $\Delta$ denotes the change across the timestep of the fingertip distance to the centroid of the object, $\Delta^t_i=||\mathsf{f_{i, t}}-\mathsf{p^C_t}||_2 - ||\mathsf{f_{i, t-1}}-\mathsf{p^C_ {t-1}}||_2$, and $\mathsf{f_i} \in \mathbb{R}^3$ denotes the position of the $i$-th fingertip, and $p_t^C$ denotes the position of the centroid of the object. 


Finally, we define a penalty on the movement of each finger: $r_v = \sum_{i=1}^3 ||\dot{\mathsf{f_i}}||_2^2$ where $\dot{\mathsf{f_i}}$ denotes the velocity of the $i$'th fingertip in the global frame.

 \noindent \textbf{4. Total Reward} Our total reward is defined as:
\vspace{-4pt}
\begin{multline*}
R(s, a) = w_{f} \times r_f \times  \mathbf{I}(t \leq \texttt{N}_v)
+ w_{v} \times r_v
+ w_{o} \times r_o
\end{multline*}
\vspace{-15pt}

where $w_{f}=-750$, $w_{v}=-0.5$ and $w_{o}=40$ are the weights of each reward component, determined through search over numerous training runs. We also found in initial experimentation that the curriculum reducing the weight of $r_f$ reward to $0$ after $\texttt{N}_v=5e7$ timesteps was needed in order to allow the robot to perform ungrasping needed to facilitate reorientation which is learned later in training. However, having the reward term during the initial phases of training sped up learning by encouraging the robot to interact with the object. In the kernel $\mathcal K$ we use $a=30$ and $b=2$ which provided a good balance between learning behaviour early in training and good accuracy later in training.



\begin{figure}[!t]
\centering
\centering
  \includegraphics[width=\linewidth]{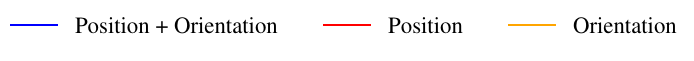}  
     \includegraphics[width=0.7\linewidth]{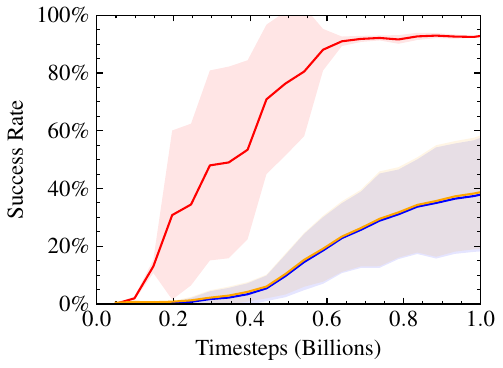}  
    \caption{
              Training curves on a reward function similar to prior work~\cite{trifinger-benchmarking, rrc-submission-chen} for the setting with DR. We take the average of 5 seeds; the shaded areas show standard deviation, noting that curves for \texttt{Orientation} and \texttt{Position+Orientation} overlap during training. It is worth noting that the nature of the reward makes it very difficult for the policy to optimize, particularly achieving an orientation goal.
    }   \label{fig:exp1}
\end{figure}

\subsection{Simulation Environment}
\label{sec:environment}

\noindent \textbf{1. Choice of Simulator} We train on the IsaacGym simulator \citep{makoviychuk2021isaac}, a simulation environment tailored towards allowing policy learning with a high sampling rate by parallising physics on a single GPU (>50K samples/sec in policy inference on Tesla V100 and around 100K samples/sec on RTX 3090). A high sampling rate is essential to learn complex dynamic robotics tasks quickly \citep{makoviychuk2021isaac, brax2021github}, and the ability to perform training and inference on desktop-level systems is important for enabling other researchers to build on our work and use it for in-hand manipulation.

\begin{figure*}[t]
\begin{minipage}[c]{0.69\textwidth}
  \centering
  \includegraphics[width=\linewidth]{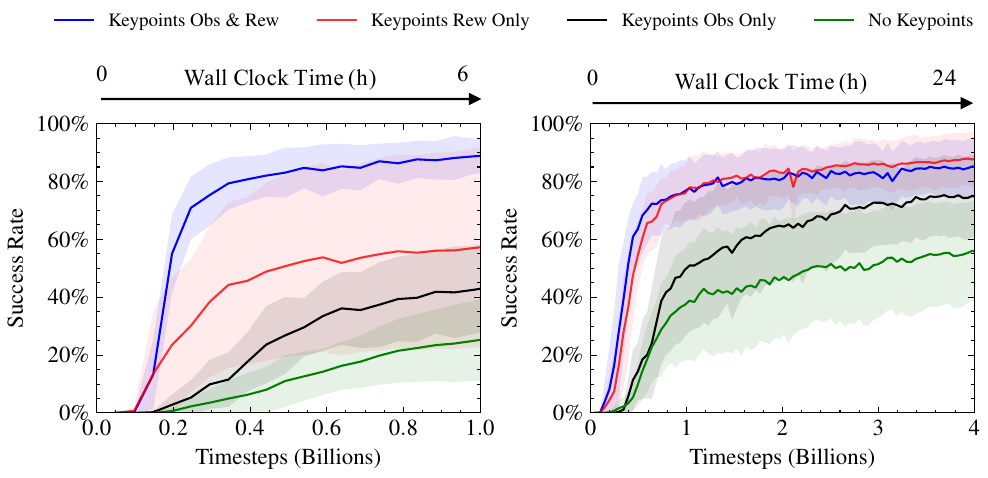}
  \centerline{
    \makebox[0.6\linewidth][c] { \footnotesize{(a) No DR.} }
    \hfill
    \makebox[0.4\linewidth][c] { \footnotesize{(b) With DR.} }
    }
\end{minipage}
\begin{minipage}[c]{0.25\textwidth}
\centering
\caption{\small
Success Rate over the course of training without and with domain randomization. Each curve is the average of 5 seeds; the shaded areas show standard deviation. Note that training without DR is shown to 1B steps to verify performance; use of DR didn't have a large impact on simulation success rates after initial training.}
\label{fig:exp2}
  \end{minipage}
\end{figure*}

\noindent \textbf{2. Domain Randomisation} Domain Randomization (DR) is a method for improving the robustness of policies for sim-to-real transfer \citep{openai-dr,Jason:ICRA:2018,mandlekar2017arpl}. In Domain Randomisation, simulator parameters are sampled from a distribution, $\xi \sim P(\mathbf{\Xi})$ in order to modify the physics behaviour of the simulated environment. If the real-world physics parameters $\xi$ are within the support of the distribution $P(\mathbf{\Xi})$, then a policy which successfully achieves the task in simulation will be able to perform comparably in the real world.

We choose our Domain Randomization parameters to account for modelling errors in the environment as well as noise in sensor measurements. These parameters are listed in \tabref{tab:dr}. In addition to these randomizations, we apply random forces to the object as described in \citep{openai-sh} in order to improve the stability of grasps and represent un-modelled dynamics. We mimic the refresh rate of the camera on the real system, by repeating the observation of the keypoints for 5 frames (See \secref{sec:environment}). To mimic possible extra camera latency, with $3\%$ probability, we repeat the camera-based object-pose observations for subsequent rounds of policy to mimic dropped frames from the tracker. Up-to-date proprioceptive data is provided to allow the policy to take advantage of the high-frequency and more reliable encoder information available on the real system.

\vspace{-5pt}
\section{Experiments}
\label{sec:experiments}
\vspace{-5pt}


In our experiments, we aim to answer the following four questions pertaining to learning a robust policy for this task, as well as evaluating how well it transfers to the real world:
\begin{enumerate}[noitemsep, wide, labelwidth=!, labelindent=0pt]
    \item How well does our system learn 6-DoF manipulation with a reward function based on prior works?
    \item How does performance change when we use a task-appropriate representation - keypoints - for reward computation and policy input in the 6-DoF reposing task?
    \item Is our system robust to sensor noise and varying environment parameters, and to changes in object morphology?
    \item How well do our policies, trained entirely in simulation, transfer to the real TriFinger system?
\end{enumerate}


\vspace{-5pt}
\subsection{Experiment 1: Training}
\label{sec:initialtrainingexp}



\noindent \textbf{1. Success Criterion} The aim in our 6-DoF manipulation task is to get the position and orientation of the object to a specified goal position and orientation. We define our metric for `success' in this task as getting the position within 2\,cm, and orientation within 0.4 rad (22\textdegree) of the target goal pose as used in \citep{openai-sh}; comparable to mean results obtained in \citep{trifinger-benchmarking}. Following previous works dealing with similar tasks \citep{openai-sh, openai-rubiks, causalworld}, we apply a reward based on the position and orientation components of error individually.

\noindent \textbf{2. Alternative Reward Formulation} Following experimentation, the best candidate reward of this format was: 

\begin{equation*}
    r_o = \mathcal{K}(||\mathsf{t^C} - \mathsf{t^T}||_2) + \frac{1}{3 \times \mathsf{|d^r|}+0.01}
\end{equation*}

where $\mathsf{t^C}$ and $\mathsf{t^T}$ are the current and goal positions of the object, $d^r = 2 \times \arcsin (\min(1.0, ||\mathsf{q_{diff}}||_2)), \mathsf{q_{diff}} = \mathsf{q^C} \mathsf{(q^T)^{*}}$. $\mathcal{K}$ is the logistic kernel that takes L2 norm between the current and target object position as input, and $\mathsf{d^r}$ is the distance in radians between the current and target object orientation. We use the alternative scaling parameter $a=50$ in $\mathcal{K}$, which we found to work better in this reward formulation (see \secref{sec:reward}). We use the same weightings for each of the 3 components of the reward as in \secref{sec:reward}. For this experiment, we trained on the $6.5 \text{cm}$ cube used in the Real Robot Challenge \citep{real-robot-challenge}.

\noindent \textbf{3. Results} The results are shown in \figref{fig:exp1}. We found that while this formulation of the reward was good at allowing PPO to learn a policy to get the object to the goal, even after 1 Billion steps in an environment with no Domain Randomization it was learning very slowly to achieve the orientation goal.

\begin{figure*}[t]
\begin{minipage}[c]{0.78\textwidth}
  \centering
  \includegraphics[width=\linewidth]{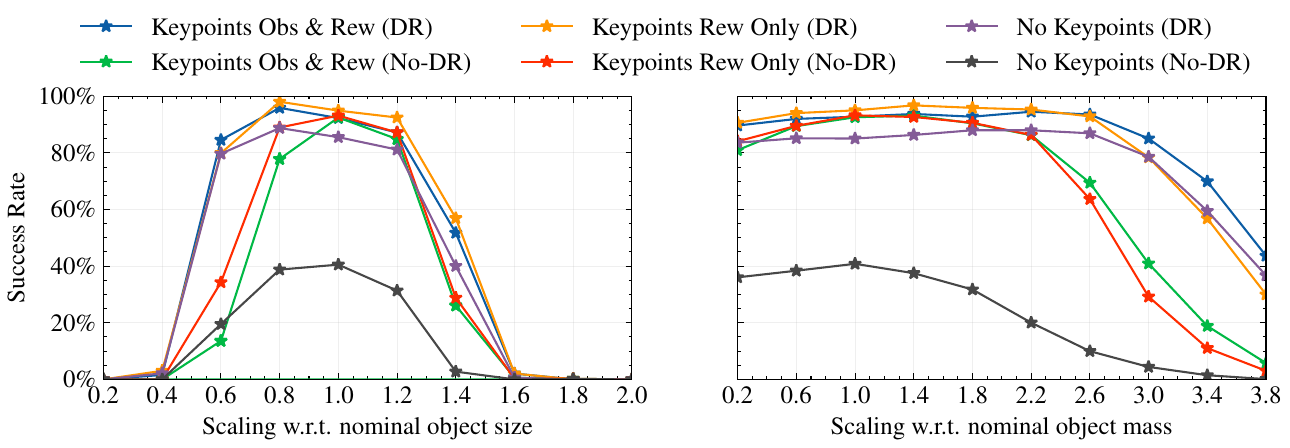}
\end{minipage}
\begin{minipage}[c]{0.21\textwidth}
\centering
\caption{We show the robustness to varying the object parameters outside the DR range it was trained for. In the evaluations, all the other DR is turned off to ensure a controlled setting. Each success is evaluated over 1024 runs with random goal and object initialization.}
\label{fig:dr_experiments_obj}
  \end{minipage}
\end{figure*}

\begin{figure*}[ht]
  \centering
  \includegraphics[width=0.9\textwidth]{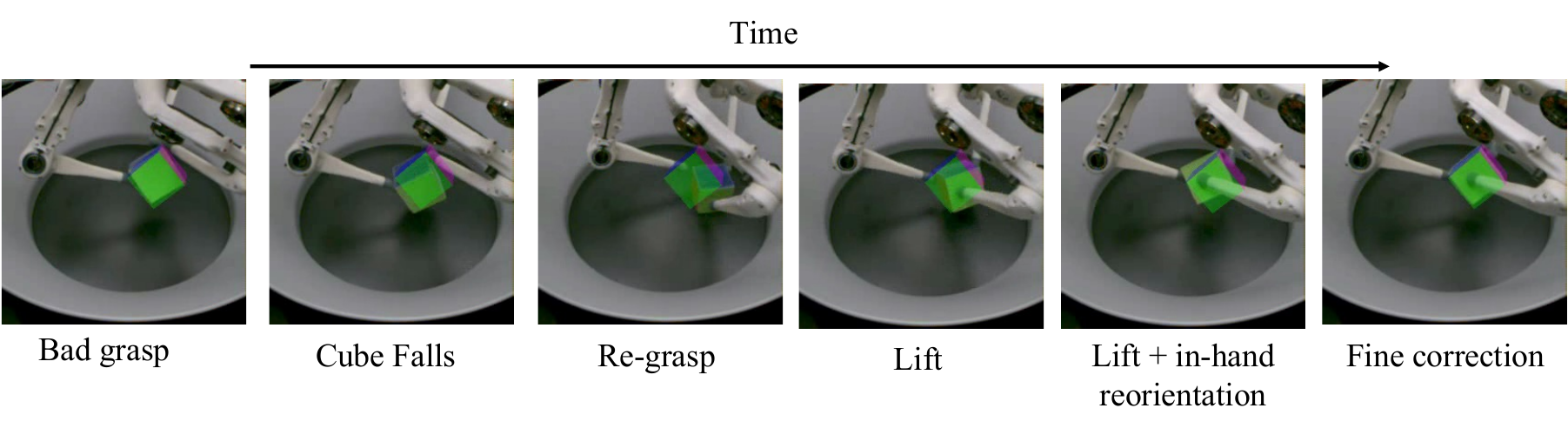}
  \caption{The use of a single, continuous policy for grasping and reposing allows the policy to automatically recover from failures. For example in the sequence above the system recovers from a failure and re-grasps the cube to achieve the desired goal pose.}
  \label{fig:qualitative-rollout}
\end{figure*}

\vspace{-3pt}
\subsection{Experiment 2: Representation of Pose}
\label{sec:posereprexp}

\noindent \textbf{1. Comparison} The poor results in Experiment 1 (\secref{sec:initialtrainingexp}) lead us to search for alternative representations of object pose in the calculation of the reward and policy observations; these are described in \secref{sec:poserepr} and \secref{sec:reward}.
We compared our method of using keypoints to represent the object pose and using positions and quaternions in two ways: firstly, using it as the policy input as compared to a position and quaternion representation, and secondly, using it to calculate the reward as compared to a reward based on the linear and angular rotational distances individually.

\noindent \textbf{2. Experimental Setup} For the observations, in order to provide a fair comparison between position/quaternion and keypoints as policy input, observation noise and delays are applied in the same manner (by applying them in the position and quaternion space before transforming to keypoints, as noted in \secref{sec:environment}). Also note that both representations only rely on the spatial pose information and fixed size of the object to compute. The pose of the object is represented with a 7-dim vector involving translation and quaternion ($\mathsf{t}$,  $\mathsf{q}$). The position and quaternion of the goal pose are provided as input to the actor and critic, replacing the keypoints in Tables \ref{table:policyobs} and \ref{table:valueobs}.

For the reward, in order to provide a fair comparison to the keypoints reward, as mentioned previously, many hours were spent tuning the kernels and parameters used in the translation based reward, described in Experiment 1. In comparison, little effort was spent tuning the keypoints function, with only one tweak to the weightings in the logistic kernel, showing the relative simplicity of working with this formulation. For this experiment, we trained on the $6.5 \text{cm}$ cube used in the Real Robot Challenge \citep{real-robot-challenge}, with keypoints placed at the bounding box (in this case the corners of the cube).

\noindent \textbf{3. Results} \figref{fig:exp2} shows the results of training, with both timesteps and wall-clock time. In the curve without any Domain Randomization, we trained for 1 billion steps over the course of 6 hours on a single GPU. Using keypoints in observations and the reward function performs the best of the four policies, also exhibiting a low variance among seeds.

When Domain Randomization is applied, the two curves with a keypoints-based reward are far better in terms of success rate at the end of training and in terms of convergence rate; however, in this case having keypoint observations seems to matter somewhat less. This is perhaps due to the longer training (4b steps \& 24 hours on a single GPU) overwhelming the inductive bias introduced by using keypoints as representations. However, using keypoints to compute the reward provided a large benefit in both cases, showing the improvement caused by calculating the reward in Euclidean space rather than mixing linear and angular displacements through addition.


\begin{figure*}[t]
\centering
  \includegraphics[width=0.9\linewidth]{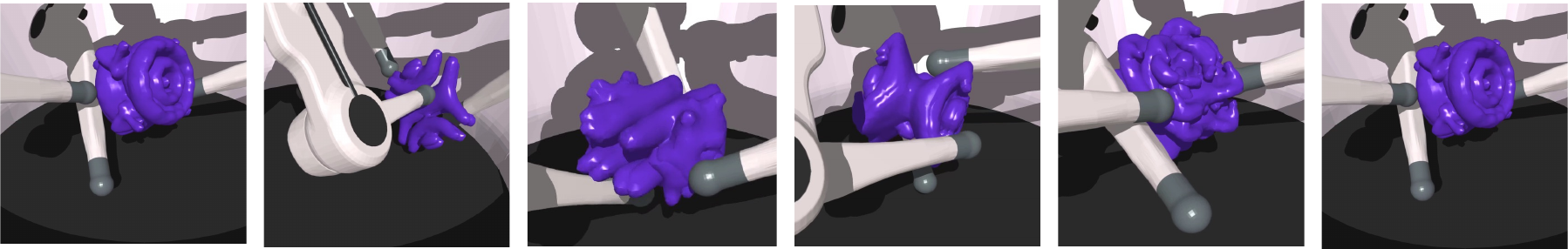}  
    \caption{
       A selection of the trifinger manipulating various different objects from the EGAD dataset. The system achieves a diversity of emergent grasps enabling it to manipulate a variety of object morphologies to 6-DoF target poses. See the website (\url{https://s2r2-ig.github.io}) for more videos of object manipulation.
    }   \label{fig:different-objects}
\end{figure*}

\vspace{-6pt}
\subsection{Experiment 3: Robustness of Policies in Simulation}
\vspace{-5pt}

\noindent \textbf{1. Impact of Varying Physics Parameters} In order to investigate the impact that Domain Randomization (see \secref{sec:environment}) has on the robustness of policies of a hand in this configuration, we ran experiments by varying parameters outside of the normal domain randomization ranges in simulation. \figref{fig:dr_experiments_obj} shows the results. We find that, despite only being randomized initially within a range of 0.97-1.03x nominal size, our policies with Domain Randomization achieve over an 80\% success rate even with a scale of 0.6 and 1.2x nominal size, while those without DR have a success rate that drops off much more quickly outside the normal range. We find similar results when scaling the object mass relative to the nominal range, however in this case we find that the policies using keypoints-based reward even without DR is much more robust at masses 3x nominal.

\noindent \textbf{2. Other objects \& Fine-Tuning} We tested the change in performance resulting from finetuning the policy (O-KP+R-KP trained with DR on the $6.5\text{cm}^3$ cube) on 100 randomly-selected objects from the EGAD dataset, and on a selecion of cuboids with random side lengths between 2-8 cm. We found an improvement in policy performance resulting from finetuning on 100 randomly selected EGAD objects, on both this same set of objects and test objects (see Table \ref{tab:finetuning-objects}). The gains were even bigger when fine-tuning on cuboids of different scales, suggesting that scale is a more important consideration than shape for generalisation.





\begin{table}[t]
\centering
\resizebox{0.9\linewidth}{!}{%
\begin{tabular}{l|c|c} 
    \toprule
    \rowcolor[HTML]{CBCEFB}
    Object & Cube Policy & Finetuned  \\
    \midrule
                            Cube 6.5cm$^3$       & 92.1\% & - \\
    \rowcolor[HTML]{EFEFEF} EGAD (train) & 84.9 \% & 87.2 \%  \\
                                EGAD (test)  & 85.2 \% & 86.4 \% \\
    \rowcolor[HTML]{EFEFEF} Different Sized Cuboids & 46.8 \% & 72.9 \%  \\

    \bottomrule
\end{tabular}
}
\caption{\textbf{Fine-tuning performance.}  We tested the zero-shot performance on the EGAD dataset and a number of differently sized cuboids to measure the robustness to differing object morphologies and scales, respectively. Numbers calculated from N=1024 trials in simulation. We disabled environment, observation \& action randomizations.}
\label{tab:finetuning-objects}
\end{table}

\vspace{-3pt}
\subsection{Experiment 4: Simulation to Remote Real Robot Transfer}
\vspace{-3pt}

\begin{figure}[!t]
\centering
  \includegraphics[width=0.8\linewidth]{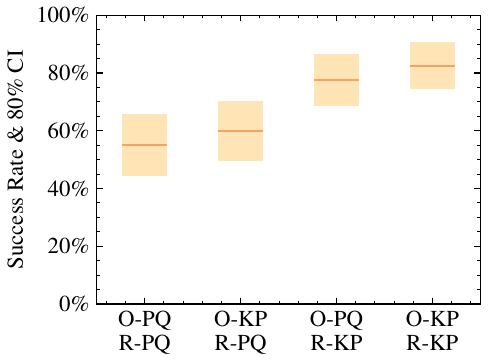}  
    \caption{
       Success Rate on the real robot plotted for different trained agents. O-PQ and O-KP stand for position+quaternion and keypoints observations respectively, and R-PQ and R-KP stand for linear+angular and keypoints based rewards respectively, as discussed in \secref{sec:posereprexp}. Each mean made of N=40 trials and error bars calculated based on an 80\% confidence interval.
    }   \label{fig:sim2real}
\end{figure}

\noindent \textbf{1. Experimental Setup} We ran experiments on the real robot to determine the success rate of the policies trained with Domain Randomization under the metric defined in \secref{fig:exp1}. We performed $N=40$ trials for each policy on the task setup described in \secref{sec:environment}; the results for each of the four ablations on keypoints which we tested are shown in \figref{fig:sim2real}.

\noindent \textbf{2. Results} Out of the four models discussed in \secref{sec:posereprexp}, the best policy achieved a success rate of 82.5\%. This was achieved with the use of keypoints used in observations of the policy as well as the reward function during training (O-KP+R-KP).    The policy using position+quaternion representations but with a reward calculated with keypoints (O-PQ+R-KP) achieved a 77.5\% success rate. These first two policies were well within each others' confidence intervals. This is likely due to the impact of the better representation of keypoints being mitigated somewhat after 4 Billion steps of training, as discussed in \secref{sec:posereprexp}. In contrast, neither of the policies trained using the position \& quaternion based reward achieved good success rates, with the policy using keypoints-based observations (O-KP+R-PQ) achieving only a 60\% success rate while the one with position and quaternion observations (O-PQ+R-PQ) only achieved a 55\% success rate. These results show the importance of having a reward function which effectively balances learning to achieve the goal in $\mathbb R^3$ and $SO(3)$ in order to have policies with a high success rate in simulation, and thus a high corresponding success rate after real robot transfer.




\noindent \textbf{3. Qualitative Behaviour} We noticed a variety of emergent behaviours used to achieve sub-goals within the overall object-reposing task. We display some of these in the panel in Figures \ref{fig:teaser} and \ref{fig:qualitative-rollout}. The most prominent of these is "dropping and regrasping". In this maneuver, the robot learns to drop the cube when it is close to the correct position, re-grasp, and pick it back up. This enables the robot to get a stable grasp on the cube in the right position. The robot learns to use the motion of the object to the correct location in the arena as an opportunity to simultaneously rotate it on the ground to make achieving the correct grasp in challenging target locations far from the center of the fingers' workspace. Our policy is also robust to dropping - it can recover from a object falling out of the hand and retrieve it from the ground.

We were only able to perform these experiments with the $6.5\text{cm}^3$ cube, as the Trifinger remote inference setup only provided a single size of cube to test on. However, using an off the shelf pose detector (for example \citep{tremblay2018deep, cosypose}), sim-to-real with these is a direction that we could pursue in the future.

\vspace{-5pt}
\section{Summary}
\vspace{-5pt}

This paper emphasizes the empirical value of a systems approach to robot learning through a case study in dexterous manipulation. 
We introduced a framework for learning in-hand manipulation tasks and transferring the resulting policies to the real world. We show how RL algorithms for in-hand manipulation can benefit from using keypoints as opposed to the more ordinary angular and linear displacement-based reward and observation computation. We show that our policies are able to generalise to unseen objects, and success rates can be further improved through finetuning. 
In contrast to prior work, our system solves all of the challenges inherent in 6-DoF grasping and reposing in a single policy, simplifying the pipeline of using RL for dexterous manipulation. We provide a clear elucidation of our approach and open source checkpoints and code to allow reproducing our work.




    


{\small
\bibliographystyle{IEEEtranN}
\bibliography{trifinger}
}

\end{document}